\documentclass[runningheads]{llncs}

\usepackage{eccv}

\usepackage{eccvabbrv}

\usepackage{graphicx}
\usepackage{booktabs}
\usepackage{makecell}
\usepackage{tikz}
\usepackage{arydshln}

\usepackage[accsupp]{axessibility}  % Improves PDF readability for those with disabilities.

\usepackage{hyperref}

\usepackage{orcidlink}

\begin{document}


\title{MSLIQA: Enhancing Learning Representations for Image Quality Assessment through Multi-Scale Learning} 

\titlerunning{MSLIQA}

\author{Nasim Jamshidi Avanaki\inst{}\orcidlink{0009-0006-9127-6249} \and
Abhijay Ghildyal\inst{1}\orcidlink{0000-0003-1940-9626} \and
Nabajeet Barman\inst{2}\orcidlink{0000-0003-2587-7370} \and
Saman Zadtootaghaj\inst{2}\orcidlink{0000-0002-6028-8507}}

\authorrunning{Jamshidi Avanaki et al.}

\institute{Portland State University, OR, USA \and Sony Interactive Entertainment, UK/Germany\\
\email{n.jamshidi.avanaki@gmail.com},
\email{abhijay@pdx.edu}\\
\email{\{Nabajeet.Barman,Saman.Zadtootaghaj\}@sony.com}}

\maketitle

\begin{abstract}

  No-Reference Image Quality Assessment (NR-IQA) remains a challenging task due to the diversity of distortions and the lack of large annotated datasets. Many studies have attempted to tackle these challenges by developing more accurate NR-IQA models, often employing complex and computationally expensive networks, or by bridging the domain gap between various distortions to enhance performance on test datasets. In our work, we improve the performance of a generic lightweight NR-IQA model by introducing a novel augmentation strategy that boosts its performance by almost 28\%. This augmentation strategy enables the network to better discriminate between different distortions in various parts of the image by zooming in and out. Additionally, the inclusion of test-time augmentation further enhances performance, making our lightweight network's results comparable to the current state-of-the-art models, simply through the use of augmentations.

\end{abstract}

\section{Introduction}
\label{sec:intro}

Image Quality Assessment (IQA) is vital in Computer Vision, impacting applications like digital photography, video streaming, medical imaging, and autonomous vehicles. Accurate IQA methods are essential for optimizing image processing systems and enhancing user satisfaction. While deep learning models have achieved significant success in IQA, their complexity makes them unsuitable for lightweight applications such as real-time image editing, live video streaming, or immediate medical diagnosis. Moreover, these models often perform poorly on datasets with unknown distortions due to limitations in their training data. Thus, addressing the challenges of model complexity and ensuring robust generalization across various distortion types is crucial for the effective deployment of IQA systems in practical settings.

While subjective methods involving human observers provide the most accurate image quality assessments, they are often costly and impractical. In contrast, objective methods, including recent neural-network-based approaches, offer more accurate predictions by better mimicking the human visual system compared to traditional statistical methods. The latest NR-IQA networks leverage Transformers~\cite{yang2022maniqa, golestaneh2022no, yun2023uniqa} and typically require a large number of parameters and substantial computational resources, limiting their suitability for fast operation or deployment on low-power devices. In contrast, our method uses MobileNetV3~\cite{howard2019searching}, a much lighter CNN network, as the baseline, making it more suitable for such applications. Compared to other CNN networks used for NR-IQA~\cite{agnolucci2024arniqa, su2020blindly}, such as ResNet-50~\cite{he2016deep}, our network utilizes significantly fewer parameters. Moreover, we do not employ loss functions for relative ranking~\cite{golestaneh2022no} or self-supervised learning with contrastive loss for pre-training on unlabeled data~\cite{agnolucci2024arniqa}. These techniques are highly effective in bridging the domain gap between the degradations observed in training and test datasets used for benchmarking NR-IQA methods. Instead, we focus on achieving similar generalization through augmentations inspired by recent work on learning to zoom in and out~\cite{taesiri2024imagenet}. 

\section{Method}
\label{sec:method}

In this paper, we demonstrate that significant improvements in image representations for image quality assessment (IQA) tasks can be achieved by employing multi-scale learning and conducting inference with Test-Time Augmentation. This approach enhances performance when inferring on datasets different from the training set. Specifically, we employ a strategy similar to that of Taesiri \etal~\cite{taesiri2024imagenet}, where different levels of zoom are achieved through resize and crop operations. % while preserving the image's original aspect ratio. 

Image resizing and cropping are crucial steps in training deep neural networks for many vision tasks. However, these processes can adversely affect model performance in IQA tasks due to the associated information loss. Resizing can result in the loss of high-frequency information, harming the model's ability to differentiate between various distortions and thus reducing performance. Cropping can omit parts of the image, leading to the learning of spurious correlations between non-essential features and the corresponding labels. Additionally, when a model trained on a dataset with specific image sizes is tested on datasets with different sizes, the negative impacts of resizing and cropping become even more pronounced. This occurs because a model trained on a dataset with images of a specific size will be biased toward the pixel distribution of that size.

We propose a straightforward solution to address this issue by training our model across multiple levels of zoom, including various crop sizes and different resizing levels. Additionally, we treat these augmentation operations as distinct datasets and use a multi-task training approach. 
We train the features extracted from these crops and their various scales using multiple heads assigned to each level of zoom. Specifically, we use crop sizes and resizes ranging from 224 to 384 as well as original image as an input, treating them as separate tasks and learning them from their respective MLP-head.

Test-time data augmentation (TTA) is a technique that estimates uncertainty and enhances model predictions. It is widely used in image classification tasks and is particularly beneficial when test inputs originate from previously unseen distributions, as augmenting the data often improves a model's generalization to new datasets~\cite{taesiri2024imagenet}. Our TTA methodology is based on multi-crop augmentation and evaluation, which involves extracting patches from the original image, along with their horizontal reflections and various scales. This approach helps improve the model's performance by providing diverse perspectives of the test image.

Although these augmentations might seem counterintuitive, they prove effective in practice. When assessing image quality, we posit that humans are influenced by recognition processes such as the processing of color, depth, and shape. Therefore, assigning multiple heads to handle different zoom levels should be beneficial. Interestingly, despite our feature extractor not being as strong, we are able to achieve performance that is either better than or on par with state-of-the-art NR-IQA methods, as shown in Table~\ref{table:benchmark}.

\begin{figure}[t]
    \centering
    \includegraphics[width=1\textwidth]{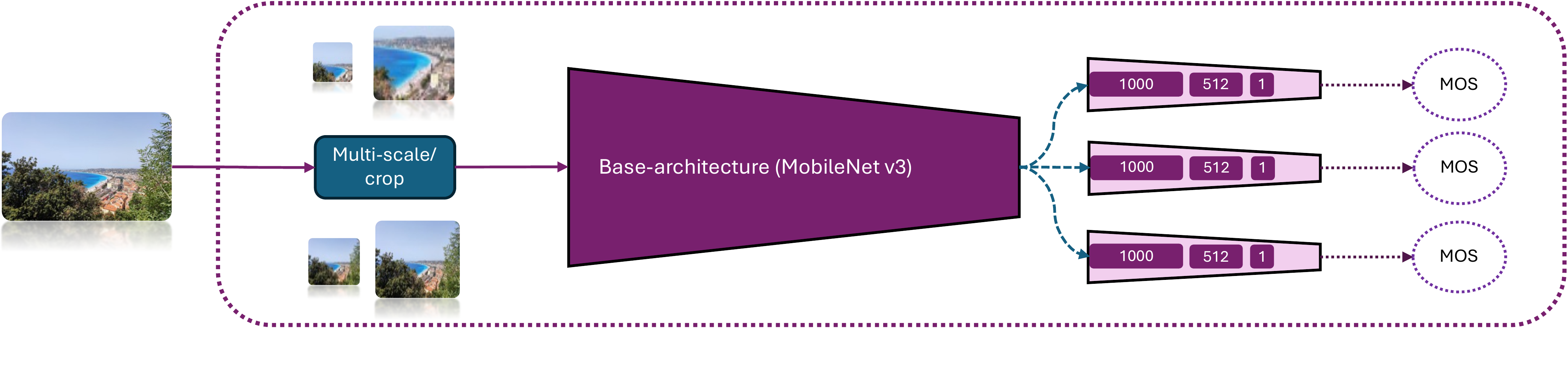}
    % \vspace{-0.2 in}
    \caption{An example of Model Structure with three MLP heads.}
    \label{fig:arch}
    % \vspace{-0.3 in}
\end{figure}

%################################################################

\section{Experiments and Results}

Figure~\ref{fig:arch} shows our multi-task learning approach assigning an MLP-head for each augmented sample. Each MLP-head consists of two fully connected layers: the first layer downscales the embedding to a vector of 512 features, and the second layer maps this 512-features vector to a single value for regression. As baseline, we train the MobileNetV3 with resizing of the input image to 224x224 with only one head. All of our models are trained using a standard loss function, which is a weighted combination of mean-squared error and PLCC loss, measured between images in the batch. 

\textbf{Training Setup:} For the Multi-cropping experiment, we used crop sizes of 224, 256, 299, 384, and no cropping (due to limited size of KADID dataset to 512x384). Similarly, for the Multi-resizing experiment, the image input was resized to 224, 256, 299, 384, and 512. For the Multi-resizing and Multi-cropping experiment, we used a combination of resizing and cropping, with resizing serving as a zooming function. Specifically, we used crop sizes of 224 and 384, as well as mixed processes: resizing to 768 followed by cropping to 384, and resizing to 512 followed by cropping to 224.

\textbf{Inference Setup:} For inference without TTA, we pass the image to the model without cropping or scaling and report the result using the best head. For inference with TTA, we use nine uniform patches of sizes 224 and 384 from three scales of the TID13 dataset (0.5x, 1x, 2x), apply transpose augmentation, and average the 108 resulting patches to determine the prediction score.

We present our ablation results in Table~\ref{table:ablation}. For this ablation study we perform cross-dataset evaluation. We train on the Kadid-10k dataset~\cite{kadid10k} with various augmentation strategies and test on the full-scale images of the TID-2013 dataset~\cite{ponomarenko2015image}. We augment the test images only for Test-Time Augmentation (TTA). As observed in Table~\ref{table:ablation}, the addition of augmentations improves the result by $\sim$28\%.

For a fair comparison, we benchmark our method against other state-of-the-art models that are also trained on the Kadid-10k dataset~\cite{kadid10k} and evaluated on the TID-2013 dataset~\cite{ponomarenko2015image}. As previously mentioned, our method is lightweight and does not utilize additional types of losses or pertaining. The results in Table~\ref{table:benchmark} show that our method outperforms most state-of-the-art methods and is comparable to ARNIQA~\cite{agnolucci2024arniqa}. 
The SRCC and PLCC values reported in Table X are sourced from the original publications. Where these values were not available, a dash (-) is used to indicate their absence.

\begin{table}[tb]
  \setlength{\tabcolsep}{3pt} 
  \centering
  \caption{Ablation study through cross-dataset evaluation. We train on the Kadid-10k dataset~\cite{kadid10k} and test on the TID-2013 dataset~\cite{ponomarenko2015image}.}
  % \vspace{-0.15 in}
  \begin{tabular}{ccccc}
    \toprule
    Multi-resizing & Multi-cropping & TTA & SRCC & PLCC  \\
    \midrule
    & & & 0.59 & 0.64 \\ 
    \checkmark & & & 0.70 & 0.71 \\
    & \checkmark & & 0.73 & 0.74 \\
    \checkmark & \checkmark & & 0.74 & 0.75 \\
    \checkmark & \checkmark & \checkmark & \textbf{0.76} & \textbf{0.78} \\
    \bottomrule
  \end{tabular}
  \label{table:ablation}
  % \vspace{-0.15 in} 
\end{table}

\begin{table}[tb]
  \setlength{\tabcolsep}{3pt}
  \centering
  \caption{Cross-dataset evaluation of various state-of-the-art models trained on KADID-10k~\cite{kadid10k} and tested on TID-2013~\cite{ponomarenko2015image}. (Best result in \textbf{bold}, 2nd-best \underline{underlined})}
  % \vspace{-0.15 in}
  \begin{tabular}{cccc}
    \toprule
    Method & Year  & SRCC & PLCC \\
    \midrule
    HyperIQA~\cite{su2020blindly} & CVPR 2020 & 0.71 & - \\
    TReS~\cite{golestaneh2022no} & WACV 2022  & 0.67 & 0.64 \\
    MANIQA~\cite{yang2022maniqa} & CVPRW 2022 & 0.75 & 0.76 \\
    Su \etal~\cite{su2023distortion} & PR 2023 & 0.69 & - \\
    YOTO~\cite{yun2023uniqa} & arXiv 2024  & 0.75 & 0.76 \\
    ARNIQA~\cite{agnolucci2024arniqa} & WACV 2024 & \textbf{0.77} & \underline{0.77} \\
    \hdashline
    Baseline (ours) & & 0.59    & 0.64 \\
    MSLIQA (ours) & & \underline{0.76}    & \textbf{0.78} \\
    \bottomrule
  \end{tabular}
  \label{table:benchmark}
  % \vspace{-0.25 in} 
\end{table}

%################################################################
\section{Conclusion}
This paper demonstrates how incorporating learned zooming data and cropping augmentations during training, combined with Test-Time Augmentation (TTA), can improve an image encoder model's generalization on the IQA task. We show that our lightweight model, enhanced with these additional augmentations, is more accurate and effective in test scenarios. Moreover, our smaller image encoder backbone can compete with models pre-trained on large datasets. In the future, we plan to study this augmentation with various backbone networks, evaluate results on multiple datasets with both real and synthetic distortions, scale the study with additional metrics and settings, compare model sizes and runtime, and assess robustness.

% \clearpage  
\bibliographystyle{splncs04}
\bibliography{main}
\end{document}